\pdfoutput=1

\documentclass[11pt]{article}

\usepackage[]{EMNLP2022}

\usepackage{times}
\usepackage{latexsym}
\usepackage{times}
\usepackage{latexsym}
\usepackage{graphicx}
\usepackage{lipsum}

\usepackage{float}
\usepackage{algorithm}
\usepackage{algorithmic}
\usepackage{placeins}
\usepackage{subfigure}
\usepackage{booktabs}
\usepackage{cleveref}
\usepackage{multirow}
\usepackage{xcolor, soul}
\definecolor{green}{HTML}{C2D8C8}
\definecolor{purp}{HTML}{D2C2D8}
\definecolor{red}{HTML}{F42C14}
\definecolor{dark-red}{HTML}{DE3163}
\sethlcolor{red}

\def\fbow{\,$\mathcal{F}_{\,\scriptsize\textsc{BoW}}$\,}

\usepackage[T1]{fontenc}

\usepackage[utf8]{inputenc}

\usepackage{microtype}

\usepackage{inconsolata}

\title{Looking at the Overlooked: \\ An Analysis on the Word-Overlap Bias in Natural Language Inference}

\author{Sara Rajaee$^1$, Yadollah Yaghoobzadeh$^2$, \and Mohammad Taher Pilehvar$^3$\\
  $^1$ University of Amsterdam, Netherlands \\
  $^2$ University of Tehran, Iran \\
  $^3$ Tehran Institute for Advanced Studies, Khatam University, Iran \\
  \texttt{s.rajaee@uva.nl}\\
  \texttt{y.yaghoobzadeh@ut.ac.ir}\\
  \texttt{mp792@cam.ac.uk}}
  
\begin{document}
\maketitle
\begin{abstract}
It has been shown that NLI models are usually biased with respect to the word-overlap between premise and hypothesis; they take this feature as a primary cue for predicting the entailment label. 
In this paper, we focus on an overlooked aspect of the overlap bias in NLI models: the \textit{reverse} word-overlap bias. 
Our experimental results demonstrate that current NLI models are highly biased towards the non-entailment label on instances with low overlap, and the existing debiasing methods, which are reportedly successful on existing challenge datasets, are generally ineffective in addressing this category of bias.
We investigate the reasons for the emergence of the overlap bias and the role of minority examples in its mitigation.
For the former, we find that the word-overlap bias does not stem from pre-training, and for the latter, we observe that in contrast to the accepted assumption, eliminating minority examples does not affect the generalizability of debiasing methods with respect to the overlap bias.
All the code and relevant data are available at:
{\url{https://github.com/sara-rajaee/reverse_bias}}

\end{abstract}

\begin{figure}[h!]
    \centering
    \includegraphics[scale=0.55]{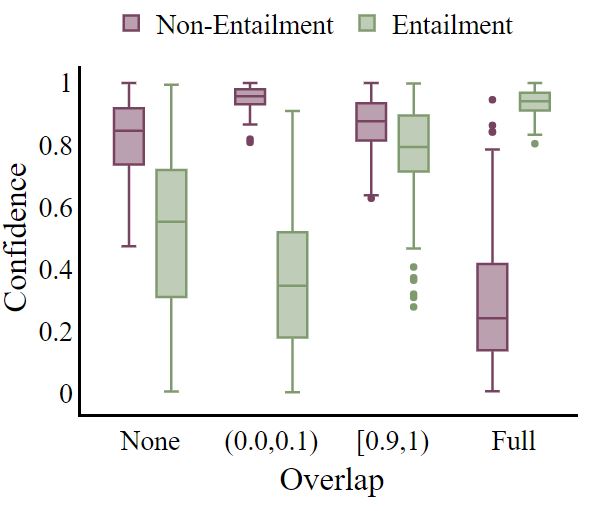}
    \caption{NLI model's confidence on a randomly sampled subset of instances from the SNLI dataset across four different degrees of word overlap between premise and hypothesis. BERT is biased towards the entailment label on instances with full overlap (denoted by the huge confidence gap with the non-entailment label). On the contrary, a \textit{reverse bias} is seen for low and non-overlapping instances, with a significant confidence lead on the non-entailment label.}
    \label{fig:reverse-bias}
\end{figure}

\section{Introduction}

Natural Language Inference (NLI) is one of the most commonly used NLP tasks, particularly in the scope of evaluating models for their language understanding capabilities.
Since their emergence, pre-trained language models (PLMs) have been highly successful on standard NLI datasets, such as the Multi-Genre Natural Language Inference \cite[MultiNLI]{williams-etal-2018-broad}. 
However, recent analytical studies have revealed that their success is partly due to their reliance on spurious correlations between superficial features of the input texts and gold labels in these datasets \cite{poliak-etal-2018-hypothesis,bhargava-etal-2021-generalization}. 
As a result, performance usually drops on out-of-distribution datasets where such correlations do not hold. 
Several proposals have been put forth to enhance the robustness of models to the known and unknown biases and improve performance on the so-called challenging datasets \cite{stacey-etal-2020-avoiding,utama-etal-2020-mind,asael-etal-2022-generative}. 


One of the well-known dataset biases in NLI models is the spurious correlation of the \textit{entailment} label and high word-overlap between premise and hypothesis. 
A number of challenging sets are designed to showcase the tendency of PLMs to predict entailment for most such cases.
HANS \cite{mccoy-etal-2019-right} is arguably the most widely used dataset in this group.
Constructed based on human-made linguistic patterns, the dataset focuses on high-overlapping samples, the non-entailment subset of which is deemed as challenging for NLI models.
Most current debiasing methods have considered the word-overlap bias as one of their main targets and have shown substantial improvements on HANS \cite{mendelson-belinkov-2021-debiasing,min-etal-2020-syntactic}.

\begin{table*}[ht!]
    \centering
    \scalebox{0.85}{
    \begin{tabular}{lp{13cm}l}
    \toprule
         \bf{Overlap} & \bf{Sample} & \bf{Label}  \\
         \midrule
         \multirow{4}{*}{Full ($1.0$)}    & \par P: A little kid in blue is sledding down a snowy hill.       \par H: A little kid in blue sledding.      & \multirow{2}{*}{Entailment} \\
         \cmidrule(lr){2-3}
                         & \par P: The young lady is giving the old man a hug.       
                           \par H: The young man is giving the old man a hug.        & \multirow{2}{*}{Non-Entailment} \\
         \midrule
         \multirow{2}{*}{$\frac{12}{13}=0.923$}    &  \par P: A woman in a blue shirt and green hat looks up at the camera.                 \par H: A woman \colorbox{purp}{wearing} a blue shirt and green hat looks at the camera                     & \multirow{2}{*}{Entailment} \\
         \cmidrule(lr){2-3}
        \multirow{2}{*}{$\frac{11}{12}=0.917$}         &  \par P: Two men in wheelchairs are reaching in the air for a basketball.      \par H: Two \colorbox{purp}{women} in wheelchairs are reaching in the air for a basketball.         & \multirow{2}{*}{Non-Entailment} \\
         \midrule
         \multirow{4}{*}{$\frac{1}{14}=0.071$}     &  \par P: Several young people sit at \colorbox{green}{a} table playing poker.
                    \par H: Youthful Human beings are gathered around \colorbox{green}{a} flat surface to play a card game.
                    & \multirow{2}{*}{Entailment} \\
                    \cmidrule(lr){2-3}
        \multirow{4}{*}{$\frac{1}{11}=0.091$}       &  \par P: A blond \colorbox{green}{woman} in a white dress sits in a flowering tree while holding a white bird.
                   \par H: The \colorbox{green}{woman} beats two eggs to make breakfast for her husband.        
                   & \multirow{2}{*}{Non-Entailment} \\
         \midrule
         \multirow{4}{*}{None ($0.0$)}    &  \par P: A couple sits in the grass.       
                            \par H: People are outside.
                            & \multirow{2}{*}{Entailment} \\
                            \cmidrule(lr){2-3}
                         &  \par P: An older women tending to a garden.
                            \par H: The lady is cooking dinner.        
                            & \multirow{2}{*}{Non-Entailment} \\
         
    \bottomrule
    \end{tabular}
    }
    \caption{NLI examples with different degrees of word-overlap (between premise and hypothesis), where the overlap is the ratio of hypothesis words that are shared with the premise. The highlighted words are the common (in green) or different (in purple) words (the samples are picked to reflect extreme cases across the word-overlap spectrum).}
    \label{tab:my_label}
\end{table*}

In this paper, we revisit the word-overlap bias in NLI and the effectiveness of existing debiasing techniques. 
Despite the popularity of this type of bias, we find that some of its aspects are generally ignored in the research community. 
If we consider word-overlap as a feature with values ranging from no to full overlap, and NLI task with two labels of entailment and non-entailment, we show that there are other kinds of spurious correlation than the popular high word-overlap and entailment. 
Specifically, as it is shown in Figure \ref{fig:reverse-bias}, we see a clear bias towards non-entailment for the low and no word-overlap values (denoted by the high performance on the non-entailment label, which comes at the price of reduced performance on the entailment class).
We will refer to this type of bias as \textit{reverse} word-overlap throughout the paper.

Through a set of experiments, we demonstrate that the overlooked reverse word-overlap bias exists in popular NLI datasets, such as MNLI and SNLI, as well as in the predictions of PLMs. Moreover, our results suggest that while existing debiasing methods can mitigate the overlap bias in NLI models to some extent, they are ineffective in resolving the reverse bias.


Moreover, we analyze how NLI models employ minority instances to enhance their generalization. 
Focusing on the forgettable debiasing method \cite{yaghoobzadeh-etal-2021-increasing}, we realize that eliminating HANS-like examples and the reverse ones do not hurt the generalization noticeably. 

In search of the origin of the bias, we employ prompt-based techniques to check whether the bias stems from pre-training.
We also verify the robustness of PLMs in a few-shot learning experiment with controlled and balanced training sets. 
Our results suggest that PLMs do not exhibit any bias towards a specific label. 
Nevertheless, introducing a few samples triggers the bias toward the entailment label. 
Furthermore, balancing the training examples with respect to their word-overlap prevents the emergence of bias to some extent.

Our contributions can be summarized as follows:
\begin{itemize}
    \item 
We expand our understanding of the word-overlap bias in NLI by revealing an unexplored spurious correlation between low word-overlap and non-entailment.
   \item  We analyze how debiasing methods work for the whole spectrum of word-overlap bias, finding that they generally fail at addressing bias for the low and non-overlapping cases. 
   \item To explore the origin of word-overlap bias in PLMs, we design several new experiments showing that, even when exposed to a few training examples, PLMs get biased towards predicting entailment.
\end{itemize}

\section{Natural Language Inference}
\label{sec:nli}

In NLI, a model is provided with two input sentences, namely \textit{premise} and \textit{hypothesis}.
The task for the model is to predict whether the hypothesis is true (\textit{entailment}), false (\textit{contradiction}), or undetermined (\textit{neutral}) given the premise.

\subsection{Bias in NLI Models}

Analyzing NLI models have demonstrated that they are sensitive to the shortcuts that appear in the dataset.
Several types of bias have been investigated in the literature, including hypothesis-only prediction, spurious correlations between certain words and labels (e.g., negation words and the non-entailment label), sensitivity to the length of hypothesis, and lexical overlap between the premise and hypothesis \cite{gururangan-etal-2018-annotation,poliak-etal-2018-hypothesis,mccoy-etal-2019-right,wu-etal-2022-generating}.
Relying on these spurious features hampers the language understanding ability of NLI models, leading to poor performance on out-of-distribution datasets where such superficial correlations do not hold \cite{he-etal-2019-unlearn,mccoy-etal-2019-right}.

\paragraph{Word-Overlap Bias.} Among the detected dataset biases, word-overlap is a quite well-studied shortcut in the NLI task \cite{zhou-bansal-2020-towards,mendelson-belinkov-2021-debiasing}.
We define word-overlap ($wo$) as the ratio of words in the hypothesis ($h$) that are shared with the premise ($p$), i.e., $\frac{|h\cap p|}{|h|}$.
Table \ref{tab:my_label} shows examples of different degrees of word-overlap.

\begin{table*}
    \centering   
    \setlength{\tabcolsep}{21pt}
    \scalebox{0.85}{
    \begin{tabular}{r c c c c c }

    \toprule
                  & MNLI-dev & HANS & HANS$+$ & HANS$-$ & W\small{A}\normalsize{NLI}  \\
    \midrule
    
    \multicolumn{1}{c}{} &
    \multicolumn{5}{c}{\textbf{BERT}} \\

    \cmidrule{2-6}
    Baseline &  84.2  \small  $\pm 0.3$ & 
                63.9  \small  $\pm 1.7$ & 
                98.5  \small  $\pm 1.2$ & 
                29.3  \small  $\pm 4.6$ & 
                56.9  \small  $\pm 0.6$ \\
                
    \cmidrule{2-6}
    Long-tuning & 83.4  \small  $\pm 0.8$ & 
                  65.8  \small  $\pm 2.3$ & 
                  99.0  \small  $\pm 0.2$ &
                  32.6  \small  $\pm 4.4$ &
                  58.0  \small  $\pm 0.6$ \\
                  
    \fbow     & 82.7  \small  $\pm 0.3$ & 
                73.8  \small  $\pm 0.5$ &
                91.8  \small  $\pm 0.4$ &
                55.9  \small  $\pm 1.3$ &
                59.0  \small  $\pm 0.3$ \\
                
    PoE         & 80.0  \small  $\pm 0.8$ &
                  66.9  \small  $\pm 2.2$ &
                  71.6  \small  $\pm 3.7$ &
                  62.2  \small  $\pm 2.7$ &
                  71.6  \small  $\pm 0.7$ \\
    \midrule
    
    \multicolumn{1}{c}{} &
    \multicolumn{5}{c}{\textbf{RoBERTa}} \\
    \cmidrule{2-6}

    Baseline &  87.2  \small  $\pm 0.2$ & 
                73.3  \small  $\pm 3.4$ &
                98.5  \small  $\pm 1.0$ & 
                48.2  \small  $\pm 7.8$ &
                59.7  \small  $\pm 1.6$\\

    \cmidrule{2-6}
Long-tuning &     86.9  \small  $\pm 0.3$ &
                  73.0  \small  $\pm 1.7$ &
                  97.8  \small  $\pm 1.2$ &
                  48.2  \small  $\pm 4.2$ &
                  60.3  \small  $\pm 0.1$ \\

    \fbow     & 85.6  \small  $\pm 0.3$ &
                78.9  \small  $\pm 0.6$ &
                88.1  \small  $\pm 2.4$ &
                69.7  \small  $\pm 2.3$ &
                62.0  \small  $\pm 1.4$\\
                
    PoE      & 84.6  \small  $\pm 0.1$ &
              77.0  \small  $\pm 1.5$ &
              79.3  \small  $\pm 6.2$ &
              71.4  \small  $\pm 3.7$ &
              73.4  \small  $\pm 0.1$ \\

\bottomrule
    \end{tabular}}

    \caption{The average accuracy of the baseline models and debiasing methods on the MNLI development (matched) set as the \textit{in-distribution} and W\small{A}\normalsize{NLI} and HANS as the \textit{out-of-distribution} datasets (HANS$+$ and HANS$-$ are entailment and non-entailment subsets, respectively).}
    \label{tab:mnli}
\end{table*}


\begin{figure*}
    \centering
    \includegraphics[scale=0.2]{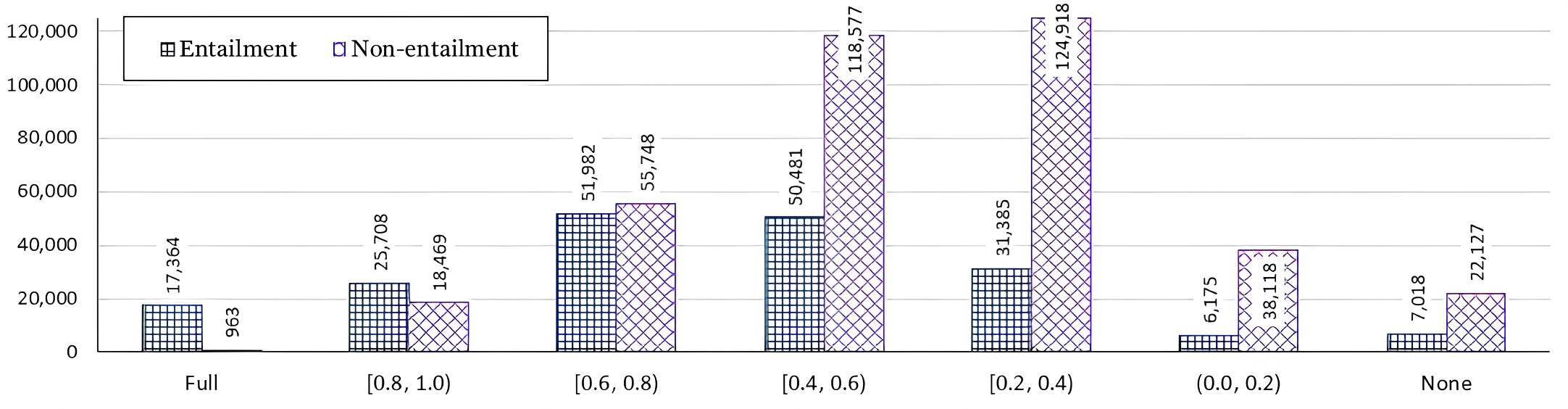}
    \caption{The distribution of instances across word-overlap bins (SNLI dataset).}
    \label{fig:probe-stat}
\end{figure*}

\subsection{Debiasing Methods}
\label{sec:debiasing}
Creating high-quality datasets without any spurious features between instances and gold labels is an arduous and expensive process \cite{gardner-etal-2021-competency}, making it inevitable for a dataset not to have biases to some extent.
Therefore, to have a robust model, it is essential to take extra steps for debiasing against dataset artifacts. 
The past few years have seen several debiasing methods \cite{karimi-mahabadi-etal-2020-end,utama-etal-2020-mind,utama-etal-2020-towards,belinkov-etal-2019-dont}. 
For our experiments, we opted for three different debiasing approaches.
We evaluate the effectiveness of these techniques in mitigating the overlap bias and its reverse. 

\paragraph{Long-tuning.}
\citet{tu-etal-2020-empirical} have shown that fine-tuning NLI models for more epochs can enhance the generalizability of LMs over challenging datasets. Following their suggestion, we fine-tuned the models for 20 epochs on the MNLI dataset.

\paragraph{Forgettable Examples.}
\citet{yaghoobzadeh-etal-2021-increasing} find minority examples without prior knowledge of the dataset artifacts. 
In the proposed method, the minority examples are considered samples that have never been learned or learned once and then forgotten by the model. Then, the already trained NLI model is fine-tuned on this subset for a few more epochs.
Following the authors' suggestion, to find the forgettable examples, we utilized a simple Siamese Bag of Words (BoW) model where the sentence representations of the premise and hypothesis are the average over their word embeddings. 

\paragraph{Product of Experts (PoE).}
In this method, a weak model is supposed to learn superficial features in the input. The weak learner's output is then used to normalize the main model's predictions on over-confident examples. Following previous studies \cite{karimi-mahabadi-etal-2020-end,DBLP:conf/iclr/Sanh0BR21}, we employed the following combination strategy for taking into account both weak learner and main model predictions:
\begin{equation}
    y = softmax(\log p_w + \log p_m )
\end{equation}
where $p_w$ and $p_m$ are the outputs of the weak learner and the main model, respectively. The robust model is trained using a cross-entropy loss function based on $y$. 
We used TinyBERT \cite{jiao-etal-2020-tinybert} as our weak learner.

\subsection{Experimental Setup}
\label{sec:setup1}

\paragraph{Datasets.} In our experiments, we opted for the Multi-Genre Natural Language Inference dataset \cite[MNLI]{williams-etal-2018-broad} for training the NLI models.
The dataset contains 433k training examples. 
Since the gold labels for the test set are not publicly available, we follow previous work and report results on the \textit{development-matched} (MNLI-dev in the tables). 
Also, following the convention in previous studies, we merge neutral and contradiction examples into the non-entailment group.
As challenging datasets, we considered HANS \cite{mccoy-etal-2019-right} and W\small{A}\normalsize{NLI} \cite{,liu-etal-2022-wanli}. 
In the former dataset, each instance is curated in a way that all words of the hypothesis are also observed in the premise, irrespective of the word order.
Previous work has shown that biased NLI models tend to perform poorly on HANS, particularly for the non-entailment class \cite{yaghoobzadeh-etal-2021-increasing}.
The latter challenging set has employed GPT-3 \cite{GPT3-NEURIPS2020} to generate high-quality instances followed by filtering done by human crowd-workers. 
Quality tests on W\small{A}\normalsize{NLI} indicate that the dataset contains fewer artifacts compared to MNLI.

\paragraph{Models.} As for PLMs, we opted for the base version of BERT and RoBERTa \cite{devlin-etal-2019-bert,liu2020roberta} and fine-tuned them for three epochs as our baselines. 
We trained the models with a learning rate of 2e-5, employing the Adam optimizer for three different random seeds. 
The batch size was set to 32 with a max length of 128. 
All the reported results are based on three random seeds.

\subsection{Results}

Table \ref{tab:mnli} shows the results for the baseline models (BERT and RoBERTa) and the three debiasing techniques on different datasets.
The bias in the baseline model is highlighted by the performance contrast across the entailment (HANS$+$) and non-entailment (HANS$-$) subsets.
As can be seen, the three debiasing methods are generally effective in softening the biased behavior, reflected by the improved performance on HANS$-$ (and, in turn, HANS), and also W\small{A}\normalsize{NLI}.

\begin{table*}
    \centering
    \setlength{\tabcolsep}{22pt}
    \scalebox{0.85}{
    \begin{tabular}{l c c c c}
      \toprule
        \multirow{2}{*}{\bf Overlap}                     & 
        \multicolumn{2}{c}{\textbf{BERT}}        &
        \multicolumn{2}{c}{\textbf{RoBERTa}}   
          \\
          \cmidrule(lr){2-3}
          \cmidrule(lr){4-5}
           & \textbf{Entailment} & \textbf{Non-Entailment} & \textbf{Entailment} & \textbf{Non-Entailment} \\ 
          \midrule
          {Full} & 99.7 \small $\pm 0.1$ & 
                          \underline{13.3} \small $\pm 1.4$ & 
                          99.7 \small $\pm 0.1$ &
                          \underline{17.6} \small $\pm 0.9$ \\
          {[0.8, 1.0)} & 92.9 \small $\pm 0.0$ & 
                               83.0 \small $\pm 1.5$ & 
                               95.9 \small $\pm 0.6$ & 
                               92.7 \small $\pm 2.4$ \\

                               \cmidrule(lr){2-5}
                               
           {[0.6, 0.8)} &      85.2 \small $\pm 0.4$ & 
                               86.2 \small $\pm 1.6$ & 
                               91.5 \small $\pm 1.4$ &
                               84.5 \small $\pm 2.8$
                                    \\

           {[0.4, 0.6)} &      74.2 \small $\pm 0.1$ & 
                               91.9 \small $\pm 1.1$ & 
                               85.8 \small $\pm 2.4$ &
                               90.2 \small $\pm 2.4$
                                    \\

           {[0.2, 0.4)} &      64.5 \small $\pm 0.6$ & 
                               95.1 \small $\pm 0.6$ &  
                               78.5 \small $\pm 2.8$ &
                               93.8 \small $\pm 1.6$
                                    \\

                               \cmidrule(lr){2-5}
                               
          {(0.0, 0.2)} & \underline{55.5} \small $\pm 1.4$ & 
                               96.7 \small $\pm 0.5$ &
                               \underline{68.6} \small $\pm 3.3$ & 
                               96.0 \small $\pm 1.2$ 
                               \\

          {None}     & {61.6} \small $\pm 1.3$ & 
                              95.2 \small $\pm 0.2$ & 
                              {77.2} \small $\pm 3.4$ &
                              93.6 \small $\pm 1.5$ \\
          \bottomrule
    \end{tabular}}

    \caption{The accuracy of the two NLI models across different overlap bins and on both subsets. The lowest numbers in each column are underlined.}
    \label{tab:reverse-performance}
\end{table*}

\section{Reverse Word-Overlap}
\label{sec:reverse-wo}

 Considering the word-overlap bias as a spectrum, the existing studies have mainly focused on a small subset of the spectrum, i.e., the case with full word-overlap and its spurious correlation with the entailment label.
 In this section, we evaluate the performance of NLI models on other areas of the spectrum and with respect to both labels (entailment and non-entailment) to broaden our insights on the robustness of these models considering the word-overlap feature.

\subsection{Probing Dataset}
\label{sec:setup}

As for this probing study, we experimented with the SNLI dataset \cite{bowman-etal-2015-large}, merging the training, development, and test sets to build a unified evaluation set. 
The set was split into seven bins based on the degree of overlap.
The statistics are reported in Figure \ref{fig:probe-stat}. 
As an example, the $[0.6,0.8)$ bin contains samples that have a word overlap (between premise and hypothesis) of greater than (and equal to) $0.6$ and less than $0.8$.





\subsection{Results}

Unless specified otherwise, the experimental setup in this experiment is the same as the one reported in Section \ref{sec:setup1}.
Table \ref{tab:reverse-performance} reports the results across different word overlap bins for both BERT and RoBERTa and for both labels. 
As expected, high contrast is observed on the full overlap subset: near-perfect NLI performance on the entailment, while poor performance on non-entailment, suggesting a strong bias towards the entailment label.
This is the conventional type of NLI bias that has been usually discussed in previous studies. 
The HANS challenging dataset is constructed based on the same type of bias. 
However, surprisingly, the results show that this biased behavior only exists for samples with full overlap.
In fact, no notable bias is observed even for the high overlap samples in the [$0.8$, $1$) bin. 
This observation further narrows down the scope of HANS as a challenging dataset and raises questions on the robustness of models developed based on the dataset.

\paragraph{Reverse bias.} Interestingly, the results
in Table \ref{tab:reverse-performance} shed light on another inherent spurious correlation that exists between NLI performance and the degree of word-overlap.
Particularly towards the non-overlap extreme, the performance drops on entailment and increases on non-entailment samples. In the (0.0, 0.2) bin, we see the largest gap: 55.5 entailment vs 96.7 non-entailment for the BERT model.
We refer to the biased behavior of NLI models on the low word-overlapping samples towards the non-entailment label as the \textit{Reverse bias}.

It is also worth mentioning that based on the proposed results, reverse bias covers a broader range of bins in comparison with the word-overlap bias.

\begin{figure*}[h!]
    \centering
    
    \includegraphics[width=
    13cm]{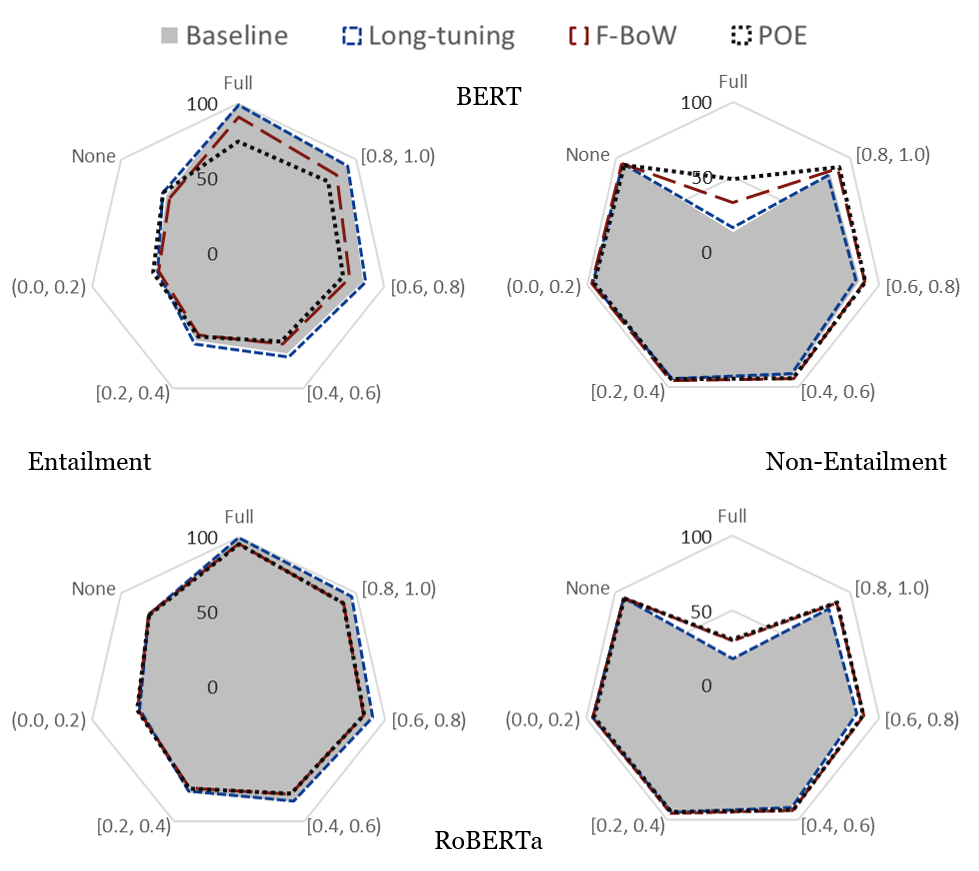}
    \caption{The performance of the baseline and the three debiasing methods across the seven word-overlap bins for both labels and for BERT and RoBERTa. Across the spectrum, the debiasing techniques seem to be effective only on samples with high (particularly full) word-overlap on the non-entailment subset and are either ineffective (or even harmful) towards the other end of the overlapping spectrum and on the entailment subset.}
    
    \label{fig:Reverse-Bias}

\end{figure*}

\subsection{Effectiveness of Debiasing Methods}

Figure \ref{fig:Reverse-Bias} shows the performance of the three debiasing methods (described in Section \ref{sec:debiasing}) across the seven bins in our word-overlap analysis. 
As can be observed, debiasing methods improve over the baseline on the full-overlap (``Full'' in Figure \ref{fig:Reverse-Bias}) and non-entailment subset, with PoE proving the most effective. 
The improvement is expected since the results on the challenging dataset, HANS, suggest the same.
This, however, comes at the price of 
reduced performance on the entailment subset, specifically in the BERT model. 

As we move toward the non-overlap end of the spectrum (``None'' in Figure \ref{fig:Reverse-Bias}), the performance gap between the entailment and non-entailment labels grows, mainly due to the drop in entailment performance. 
Interestingly, the experimental results reveal that debiasing methods are clearly ineffective in addressing the reverse bias and perform similarly to the baseline models.


\section{Analysis}
\label{sec:analysis}

\begin{figure*}
    \centering
    \subfigure[Non-Entailment]{
    \includegraphics[width=5.cm,height=4.cm]{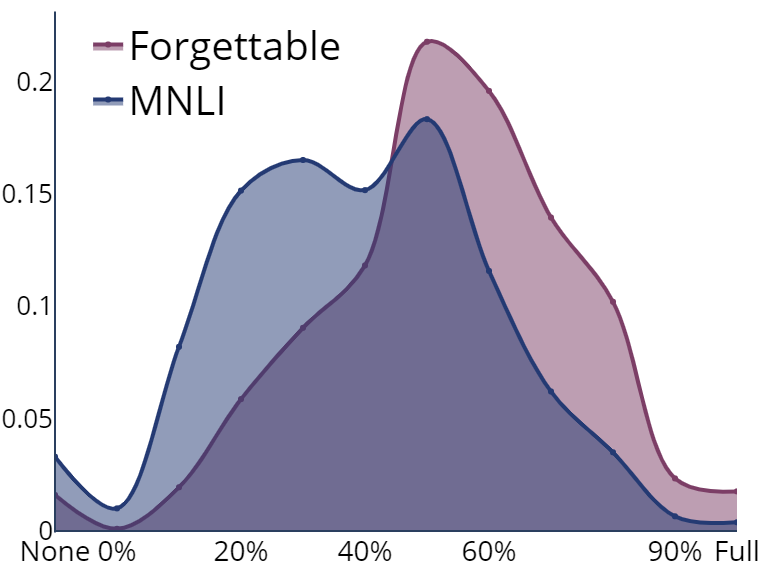}\label{fig:dist-minority}}
    \hspace{0.1\textwidth}
    \subfigure[Entailment]{
    \includegraphics[width=5.cm,height=4.cm]{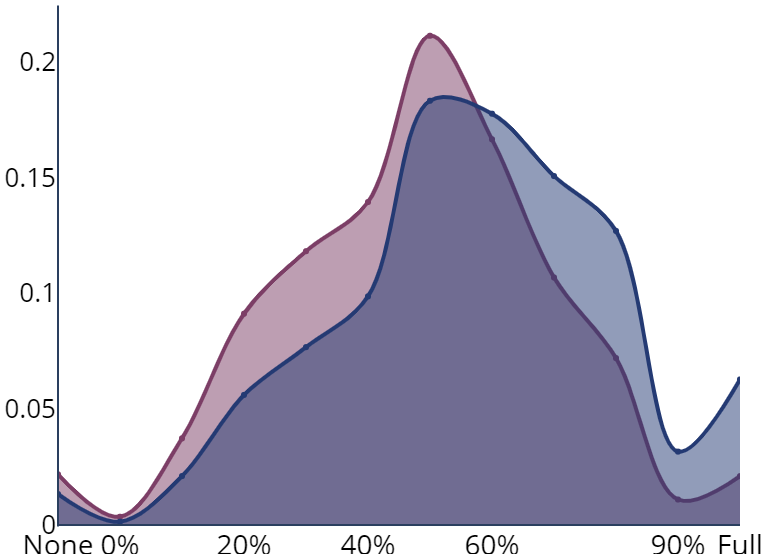} \label{fig:dist-minority-ent}}
    \caption{Normalized distribution of instances with respect to their word-overlap in the original training set of MNLI and the subset identified by \fbow. }
    \label{fig:data-dist}
\end{figure*}

\subsection{Role of Minority Examples}

In the context of word-overlap bias, the non-entailment instances that have full overlap (between premise and hypothesis) are usually referred to as \textit{minority} examples. 
\citet{tu-etal-2020-empirical} show that minority examples of the training set play a crucial role in the generalizability of language models, and eliminating them can significantly hurt performance on challenging datasets, such as HANS.
\citet{yaghoobzadeh-etal-2021-increasing} relate the forgettables with the minority examples by observing the difference in word-overlap distribution in forgettables.

We carry out a set of experiments on the \textit{forgettable} approach, where a subset of the training data is chosen for further fine-tuning of models (66$k$ in our NLI experiments for the \fbow method). 
We extend the forgettable analysis to the low word-overlap or reverse minority examples. 
We also verify the role played by minority examples in the performance of debiasing methods. 

As the first step, we compare the distribution of instances with respect to their overlap in the original training set of MNLI and its forgettable subset.
The results are shown in Figure \ref{fig:data-dist}. 
As can be seen, the forgettable subset tends to have better coverage over the minority subset than the original MNLI training set. See the right side of Figure \ref{fig:dist-minority} and the left side of Figure \ref{fig:dist-minority-ent}.

One can hypothesize that better coverage of minority examples is the reason behind the effectiveness of the forgettable approach. 
To verify this hypothesis, we eliminate several subsets from \fbow and fine-tune the NLI models with the remaining samples. 
We considered the following four settings:

\begin{itemize}
    \item \textbf{Full $-$ NEnt:} Full overlap between premise and hypothesis with the non-entailment label.
    \item \textbf{None $-$ Ent:} No overlap and entailment label.
    \item \textbf{[0.8, 1.0] $-$ NEnt:} More then  80\% overlap and non-entailment label.
    \item \textbf{[0.0, 0.2] $-$ Ent:} Less than 20\% overlap and entailment label.
\end{itemize}

The results are reported in Table \ref{tab:minority-role}. 
Interestingly, we observe that removing HANS-like examples (Full$-$NEnt), which were hypothesized to play the main role in improving performance on the challenging datasets, does not affect the performance of \fbow notably. 
The observation is consistent even for larger subsets of high-overlapping instances ([0.8, 1]$-$NEnt). 
Discarding the reverse group (low-overlapping entailment samples) yields a similar pattern. So, it can be inferred that such samples do not play the primary role in the debiasing methods' effectiveness. 

This opens up questions on how NLI models extrapolate to patterns unseen during training and how debiasing methods enhance their generalization over out-of-distribution data.
This is particularly interesting in light of observations made by \citep{tu-etal-2020-empirical} that standard training does not enable such extrapolation.
We leave further investigations in this area to future work.

\begin{table*}[h!]
    \centering   
    \setlength{\tabcolsep}{14pt}
    \scalebox{0.85}{
    \begin{tabular}{r c c c c c r}

    \toprule
              & MNLI-dev & HANS & HANS$+$ & HANS$-$ & W\small{A}\normalsize{NLI} & Eliminated \\
    \midrule          
    \multicolumn{1}{c}{} &
    \multicolumn{5}{c}{\textbf{BERT}} &
    \multicolumn{1}{c}{}\\
    
    \cmidrule{2-6}
    \it Baseline & 84.2  \small  $\pm 0.3$ & 
            63.9  \small  $\pm 1.7$ & 
            98.5  \small  $\pm 1.2$ & 
            29.3  \small  $\pm 4.6$ & 
            56.9  \small  $\pm 0.6$ \\
    
    \fbow &  82.7  \small  $\pm 0.3$ & 
                73.8  \small  $\pm 0.5$ &
                91.8  \small  $\pm 0.4$ &
                55.9  \small  $\pm 1.3$ &
                59.0  \small  $\pm 0.3$ &
                \\
    \cmidrule{2-7}
    Full~$-$~NEnt   & 82.8  \small  $\pm 0.4$ & 
                  71.7  \small  $\pm 0.9$ & 
                  93.2  \small  $\pm 0.4$ &
                  50.3  \small  $\pm 2.0$ &
                  59.4  \small  $\pm 0.5$ & 782\\
    
    [0.8, 1.0]~$-$~NEnt     & 83.2 \small  $\pm 0.2$ & 
                72.3  \small  $\pm 0.8$ &
                93.5  \small  $\pm 1.3$ &
                51.1  \small  $\pm 2.9$ &
                58.8  \small  $\pm 0.5$ & 6,350 \\
                
    [0.0, 0.2]~$-$~~~~Ent   & 82.9  \small  $\pm 0.4$ &
                  73.7  \small  $\pm 0.7$ &
                  91.9  \small  $\pm 0.8$ &
                  55.4  \small  $\pm 2.1$ &
                  59.5  \small  $\pm 0.7$ & 1,801 \\
                  
    None~$-$~~~~Ent         & 82.8  \small  $\pm 0.5$ &
                  73.8  \small  $\pm 0.8$ &
                  92.1  \small  $\pm 1.5$ &
                  55.5  \small  $\pm 3.1$ &
                  59.3  \small  $\pm 0.6$ & 482 \\
                  
    \midrule
    
     &
    \multicolumn{5}{c}{\textbf{RoBERTa}} 
    \\ 
    \cmidrule{2-6}
    
        \it Baseline &  87.2  \small  $\pm 0.2$ & 
                73.3  \small  $\pm 3.4$ &
                98.5  \small  $\pm 1.0$ & 
                48.2  \small  $\pm 7.8$ &
                59.7  \small  $\pm 1.6$\\
                
       \fbow &   85.6  \small  $\pm 0.3$ &
                78.9  \small  $\pm 0.6$ &
                88.1  \small  $\pm 2.4$ &
                69.7  \small  $\pm 2.3$ &
                62.0  \small  $\pm 1.4$ & \\
       \cmidrule{2-7}
        
        Full~$-$~NEnt   & 86.4  \small  $\pm 0.2$ &
              79.1  \small  $\pm 1.3$ &
              92.1  \small  $\pm 1.6$ &
              66.1  \small  $\pm 4.0$ &
              62.2  \small  $\pm 0.9$ &
              782 \\

        [0.8, 1.0]~$-$~NEnt     &      86.6  \small  $\pm 0.2$ &
                78.4  \small  $\pm 1.0$ &
                95.9  \small  $\pm 0.8$ &
                60.8  \small  $\pm 2.8$ &
                61.8  \small  $\pm 0.7$ & 
                6,350 \\
                
      [0.0, 0.2]~$-$~~~~Ent   &  86.1  \small  $\pm 0.2$ &
                  79.3  \small  $\pm 1.3$ &
                  89.8  \small  $\pm 1.2$ &
                  68.7  \small  $\pm 2.1$ &
                  62.3  \small  $\pm 0.9$ & 
                  1,801  \\     
                  
    None~$-$~~~~Ent         &  86.1  \small  $\pm 0.2$ &
              79.1  \small  $\pm 1.2$ &
              88.6  \small  $\pm 2.1$ &
              69.5  \small  $\pm 2.9$ &
              62.1  \small  $\pm 0.7$ & 
              482\\
    
    \bottomrule
    \end{tabular}}

    \caption{The performance of \fbow after eliminating four different subsets. \textit{Eliminated} denotes the number of eliminated examples in each setting. All the subsets tend to be in the same performance ballpark with respect to the generalizability of the model on the out-of-distribution datasets (W\small{A}\normalsize{NLI} and HANS). }
    \label{tab:minority-role}
\end{table*}

\begin{table*}
    \centering
    \setlength{\tabcolsep}{22pt}
    \scalebox{0.85}{
    \begin{tabular}{c c c c c c }
    \toprule
    \multicolumn{1}{c}{} &
    \multicolumn{5}{c}{\textbf{Baseline}} \\
    \cmidrule(lr){2-6}
         & MNLI-dev & HANS & HANS$+$ & HANS$-$ & W\small{A}\normalsize{NLI}  
             \\
    \midrule
    Zero-shot & 42.0 & 55.3 & 57.5 & 53.1 & 58.0 
    
    \\
    \midrule
    $K = ~16$  & 45.6 \small $\pm ~1.2$ &
                53.6 \small $\pm ~1.3$ &
                73.2 \small $\pm 16.5$ & 
                34.4 \small $\pm 13.9$ &
                54.7 \small $\pm ~2.3$ \\
    $K = ~32$  & 46.9 \small $\pm ~0.6$ & 
                50.8 \small $\pm ~0.8$ & 
                98.3 \small $\pm ~1.2$ &
                ~3.3 \small $\pm ~2.8$ & 
                50.1 \small $\pm ~2.2$ \\
    $K = ~64$  & 49.6 \small $\pm ~0.3$ & 
                50.3 \small $\pm ~0.3$ & 
                99.4 \small $\pm ~0.5$ & 
                ~1.1 \small $\pm ~1.1$ & 
                48.4 \small $\pm ~4.3$ \\
    $~K = 128$ & 52.7 \small $\pm ~0.9$ & 
                50.0 \small $\pm ~0.0$ & 
                99.9 \small $\pm ~0.2$ & 
                ~0.1 \small $\pm ~0.2$ & 
                45.1 \small $\pm ~0.4$ \\
    $~K = 256$ & 56.4 \small $\pm ~0.4$ & 
                50.7 \small $\pm ~0.8$ & 
                98.1 \small $\pm ~2.2$ & 
                ~3.3 \small $\pm ~3.9$ & 
                50.3 \small $\pm ~0.0$ \\
    $~K = 512$ & 61.4 \small $\pm ~1.1$ & 
                50.0 \small $\pm ~0.1$ & 
                100  \small $\pm ~0.0$ & 
                ~0.1 \small $\pm ~0.1$ & 
                46.2 \small $\pm ~2.0$ \\
                
    \midrule
    \multicolumn{1}{c}{} &    \multicolumn{5}{c}{\textbf{Balanced}}\\
    \cmidrule(lr){2-6}
    $K = 16$ & 
                44.1 \small $\pm ~0.6$ & 
                52.5 \small $\pm ~1.5$ & 
                95.6 \small $\pm ~2.6$ & 
                ~9.3 \small $\pm ~5.7$ & 
                54.3 \small $\pm ~3.2$ \\

    $K = 32$ &  45.7 \small $\pm ~1.3$ & 
                51.9 \small $\pm ~1.1$ & 
                82.2 \small $\pm 15.7$ & 
                21.5 \small $\pm 13.4$ & 
                52.0 \small $\pm ~1.3$ \\
    $K = 64$ & 45.2 \small $\pm ~1.1$ & 
                52.4 \small $\pm ~1.1$ & 
                69.8 \small $\pm ~6.0$ & 
                35.1 \small $\pm ~3.8$ &
                54.4 \small $\pm ~0.3$\\
    $~K = 128$ &48.0 \small $\pm ~0.1$ &
                51.7 \small $\pm ~0.1$ &
                95.7 \small $\pm ~5.0$ &
                ~7.7 \small $\pm ~5.2$ &
                52.8 \small $\pm ~3.3$ \\
                
    $~K = 256$&51.3 \small $\pm ~1.3$ &
                51.2 \small $\pm ~3.0$ &
                84.9 \small $\pm 15.6$ &
                17.5 \small $\pm 21.5$ &
                51.8 \small $\pm ~3.5$\\
                
    $~K = 512$&53.2 \small $\pm ~0.2$ &
                51.3 \small $\pm ~2.8$ &
                86.8 \small $\pm 10.8$ &
                15.8 \small $\pm 16.5$ &
                49.5 \small $\pm ~1.7$\\
    
    \bottomrule
    \end{tabular}}
    \caption{Zero-shot and few-shot results of prompt-based fine-tuning for BERT. While no significant bias is seen in the zero-shot setting, only with a few task-specific examples, BERT predictions are biased towards entailment (HANS$+$ vs. HANS$-$). Balancing the training set (bottom block) slightly reduces the extent of bias.}
    \label{tab:prompting}
\end{table*}

\subsection{The Origin of Word-Overlap Bias}

We conducted another experiment to see if the vulnerability of NLI models to the word-overlap feature and the reverse bias comes from pre-training or from fine-tuning on the task-specific data. 
To this end, we followed \citet{utama-etal-2021-avoiding} in evaluating pre-trained models under zero- and few-shot settings. 
To rule out the impact of fine-tuning and verify if the pre-trained model exhibits similar biases with respect to word-overlap, we evaluated BERT in a zero-shot setting by reformulating the NLI task as a masked language modeling objective.
Following previous studies \cite{schick-schutze-2021-exploiting, utama-etal-2021-avoiding}, we transformed the NLI examples using the below template:

\begin{verbatim}
    Premise ? [MASK], Hypothesis.
\end{verbatim}
    
\noindent where the \texttt{[MASK]} token denotes the gold label. 
We used a simple verbalizer with \textit{yes}, \textit{maybe}, and \textit{no} as mappings to, respectively, the entailment, neutral, and contradiction labels.

The first row of Table \ref{tab:prompting} shows the results for the zero-shot setting. 
The similar performance across HANS$-$ and HANS$+$ shows that the pre-trained BERT model does not exhibit much bias towards a specific label.
Therefore, the bias stems from the fine-tuning on the task-specific instances.
This is reflected even with as few as 16 samples in the few-shot scenario (where we have fine-tuned the prompt-based model).
As the number of training instances increases, the gap between the entailment and non-entailment samples grows. 

\paragraph{Balanced data.} We also examined the role of class imbalance in the training data on the emergence of word-overlap bias. 
For this experiment, we defined four categories based on the overlap \{Full, [0.5, 1), (0.0, 0.5), and None\} and uniformly sampled $K$ instances per label. 
The bottom block of Table \ref{tab:prompting} presents the results. It can be inferred that having a balanced training set can reduce the bias to some extent. 
Finally, the high variance on the HANS subsets suggests that the quality of training examples and word-overlap percentage between the premise and hypothesis can have a significant impact on the bias in NLI systems.

\section{Related Work}

\paragraph{Dataset biases in NLP.}

Different categories of bias have been discovered and discussed in NLP datasets. 
Earlier work has discovered that negative words are correlated with contradiction label in the SNLI dataset \cite{naik-etal-2018-stress,gururangan-etal-2018-annotation}.
Hypothesis-only \cite{gururangan-etal-2018-annotation} and word-overlap between hypothesis and premise \cite{mccoy-etal-2019-right} are other types of biases discussed in the literature of SNLI and MNLI datasets.
In particular, word overlap has also been investigated in the context of duplicate question detection on the QQP dataset  \cite{zhang-etal-2019-paws}. 
For both NLI and QQP, it has been shown that considerable spurious correlations exist between high word overlap and the entailment/duplicate label.
In this word, we focused on the word overlap bias in the NLI dataset and introduced an overlooked aspect of this bias: the correlation between low word overlap and non-entailment class.

\paragraph{Challenging sets.}
In the past few years, several challenging datasets have been introduced to study the limitations of NLP models and, in particular, pre-trained language models in learning robust features and ignoring dataset biases.
Challenging datasets for NLI include HANS \cite{mccoy-etal-2019-right}, ANLI \cite{williams-etal-2022-anlizing}, MNLI-hard \cite{gururangan-etal-2018-annotation} and
Stress-tests \cite{naik-etal-2018-stress}.
Similar datasets for other tasks include PAWS \cite{zhang-etal-2019-paws,pawsx2019emnlp}, for duplicate question detection, and FEVER-Symmetric \cite{schuster-etal-2019-towards}, for stance detection.

\paragraph{Spurious correlation.}
\citet{gardner2021competency} argue that for complex language understanding tasks, any simple feature correlation should be considered spurious, e.g., ``not'' and the contradiction label in NLI.
Spurious correlations can also be defined from the viewpoint of generalizability \citet{chang-etal-2021-robustness,yaghoobzadeh-etal-2021-increasing}. 
According to this definition, a feature is spurious if it works well only for specific examples.
The reverse word overlap feature described in this paper fits well within both definitions.
\citet{schwartz2022limitations} review several definitions for spurious correlations.

\paragraph{Debiasing methods.}
Many studies try to remove the spurious correlations or dataset biases either from the training dataset or the model.
Most debiasing approaches filter or down weight those training examples that are either easy or contain spurious correlations \cite{he-etal-2019-unlearn,karimi-mahabadi-etal-2020-end,utama-etal-2020-mind,DBLP:conf/iclr/Sanh0BR21}.
Others augment the training set with examples that violate the spurious correlations.
A mix of both these approaches has also been investigated by \citet{wu-etal-2022-generating}.
An alternative approach is to extend the fine-tuning either on all \cite{tu-etal-2020-empirical} or parts of training data \cite{yaghoobzadeh-etal-2021-increasing}.

\paragraph{Analysis of debiasing.}
Given the increasing interest in debiasing methods, there have been concerns about their widespread use. \citet{schwartz-stanovsky-2022-limitations} argue that excessive balancing prevents the models from learning anything (in particular, important world and commonsense knowledge), making it neither practical nor desired.
They suggest abstaining and interacting with the user when the contextual information is not sufficient and also focus on zero- and few-shot learning approaches instead of full fine-tuning. 
In this paper, we showed that balancing datasets should only be taken as a partial solution for eliminating spurious correlations. 
We also showed that in this context, few-shot learning might not be effective. 
\citet{mendelson-belinkov-2021-debiasing} found that debiasing methods encode more extractable information about the bias in their inner representations. 
This observation is explained in a concurrent work to ours in terms of the necessity and sufficiency of the biases \cite{joshi2022all}.
In this paper and for the word-overlap bias, we showed that our selected debiasing techniques are not robust against if we consider the whole spectrum. 

\section{Conclusions}

In this work, we uncovered an unexplored aspect of the well-known word-overlap bias in the NLI models. We showed a spurious correlation between the low overlap instances and the non-entailment label, namely the \textit{reverse} word-overlap bias. 
We demonstrated that existing debiasing methods are not effective in mitigating the reverse bias. 
%
We found that the generalization power of debiasing methods (the forgettable approach in particular) does not stem from minority examples.
We also showed that the word-overlap bias does not seem to come from the pre-training step of PLMs.
As future work, we plan to focus on designing new debiasing methods for mitigating the reverse bias for NLI and similar tasks. Also, building specific challenging sets, similar to HANS, for the reverse bias helps to expand this line of research.

\section{Acknowledgements}
We would like to acknowledge that the idea of reverse bias was initiated in discussion with Alessandro Sordoni (MSR Montreal). Also, we want to thank the anonymous reviewers for their valuable comments, which helped us in improving the paper. Sara Rajaee is funded in part by the Netherlands Organization for Scientific Research (NWO) under project number VI.C.192.080.

\section{Limitations}
In our experiments, we have focused on two popular PLMs, BERT and RoBERTa. Using more PLMs, with diversity in the objective and architecture and evaluating their robustness is one of the extendable aspects of our work. Moreover, we evaluated three debiasing methods, but this could have been expanded to more. The other susceptible aspect to improvement is creating a more high-quality dataset for analyzing the overlap bias and its reverse. We have used SNLI as our main probing set, a crowdsourcing-based dataset that contains some noisy examples, especially in minority groups. 

\bibliography{anthology,custom}
\bibliographystyle{acl_natbib}

\appendix

    
    


\end{document}